\documentclass[letterpaper]{article} 
\usepackage{aaai24}  
\usepackage{times}  
\usepackage{helvet}  
\usepackage{courier}  
\usepackage[hyphens]{url}  
\usepackage{graphicx} 
\usepackage{natbib}  
\usepackage{caption} 
\usepackage{algorithm}
\usepackage{algorithmic}
\usepackage{makecell}
\usepackage{subfigure}
\usepackage{multirow}
\usepackage{verbatim}
\usepackage{amsmath,amssymb}
\usepackage{booktabs}

\usepackage{newfloat}
\usepackage{listings}
\DeclareCaptionStyle{ruled}{labelfont=normalfont,labelsep=colon,strut=off} 
\floatstyle{ruled}
\newfloat{listing}{tb}{lst}{}
\floatname{listing}{Listing}
\title{Arbitrary-Scale Point Cloud Upsampling by Voxel-Based Network with Latent Geometric-Consistent Learning}
\author{
    Hang Du\equalcontrib,
    Xuejun Yan\equalcontrib,
    Jingjing Wang,
    Di Xie,
    Shiliang Pu
}
\affiliations{

     Hikvision Research Institute, Hangzhou, China\\
    \{duhang, yanxuejun, wangjingjing9, xiedi, pushiliang.hri\}@hikvision.com
}

\usepackage{bibentry}

\begin{document}

\maketitle

\begin{abstract}
Recently, arbitrary-scale point cloud upsampling mechanism became increasingly popular due to its efficiency and convenience for practical applications.
To achieve this, most previous approaches formulate it as a problem of surface approximation and employ point-based networks to learn surface representations.
However, learning surfaces from sparse point clouds is more challenging, and thus they often suffer from the low-fidelity geometry approximation.
To address it, we propose an arbitrary-scale Point cloud Upsampling framework using Voxel-based Network (\textbf{PU-VoxelNet}).
Thanks to the completeness and regularity inherited from the voxel representation, voxel-based networks are capable of providing predefined grid space to approximate 3D surface, and an arbitrary number of points can be reconstructed according to the predicted density distribution within each grid cell.
However, we investigate the inaccurate grid sampling caused by imprecise density predictions.
To address this issue, a density-guided grid resampling method is developed to generate high-fidelity points while effectively avoiding sampling outliers.
Further, to improve the fine-grained details, we present an auxiliary training supervision to enforce the latent geometric consistency among local surface patches.
Extensive experiments indicate the proposed approach outperforms the state-of-the-art approaches not only in terms of fixed upsampling rates but also for arbitrary-scale upsampling. The code is available at
https://github.com/hikvision-research/3DVision

\end{abstract}

\section{Introduction}
3D point clouds are widely used in many real-world applications~\cite{Qi2017PointNetDH,Yuan2018PCNPC,Li2018PointCNNCO,Guo2021DeepLF,PointAttN}.
However, raw point clouds captured by 3D scanners are often sparse and noisy.
Point cloud upsampling is a necessary procedure to obtain dense and high-fidelity point, enabling to provide more geometric and semantic information for downstream tasks, such as 3D classification~\cite{Qi2017PointNetDL,Guo2021PCTPC},  rendering~\cite{Chang2023PointersectNR}, and  reconstruction~\cite{ma2021neural}.

\begin{figure}[t]
    \centering
    \includegraphics[height=5.4cm]{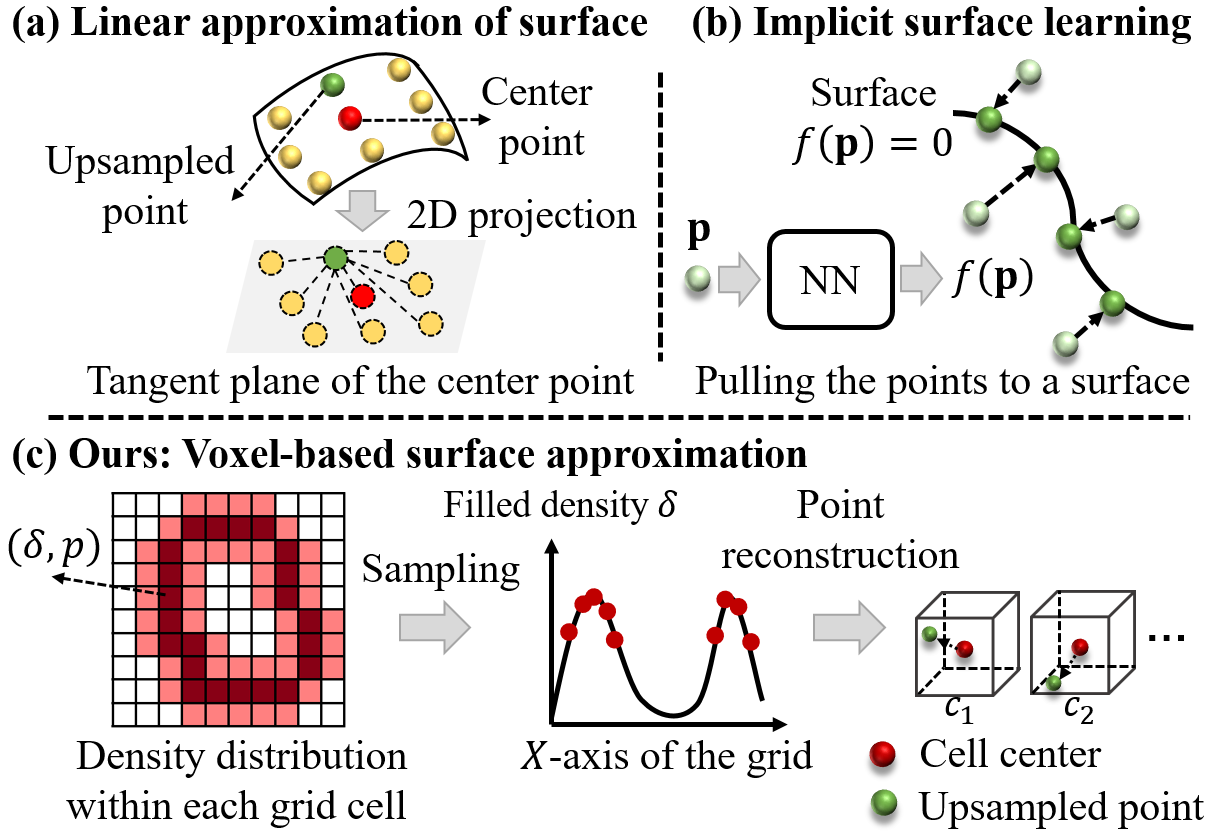}
    \caption{Differences between previous representative methods~\cite{Qian2021DeepMU,SelfPCU} and our voxel-based surface approximation for arbitrary-scale point cloud upsampling.
    Resorting to the regular structure of voxel grids, the surface patch is approximated as a density distribution of points within each grid cell, and then we can reconstruct an arbitrary number of points from the density predictions. }
    \label{figure1}
\end{figure}

Most learning-based upsampling approaches~\cite{Yu2018PUNetPC,Yu2018ECNetAE,Wang2019PatchBasedP3,Li2019PUGANAP,Qian2020PUGeoNetAG,Li2021PointCU,Qian2021PUGCNPC} merely support fixed rates, since their network architectures are specifically designed to be coupled with the upsampling rate.
After one-time training, they can only be used for a fixed upsampling rate, which are inefficient and inconvenient for practical applications.
To fulfill upsampling with varying rates, they need to build multiple networks trained at different rates or repeat running the fixed-scale model multiple times, thus increasing the storage cost and computation burden~\cite{SelfPCU}. So, they are not able to handle varying upsampling rates effectively.

In recent years, arbitrary-scale upsampling is more desirable for practical applications, which allows for upsampling with flexible rates in a one-time training.
To achieve this, some works~\cite{Qian2021DeepMU,LuoPUEVAAE} focus on learning linear approximation of local surface (Fig. 1(a)), and others~\cite{feng2022neural,SelfPCU,He_2023_CVPR} apply implicit surface learning to sample arbitrary points from the learned surface or push the points towards the underlying surface (Fig. 1(b)).
In general, existing methods tackle arbitrary-scale upsampling by formulating it as a surface approximation problem, and utilize point-based networks to construct surface representations.
However, learning accurate surfaces from sparse point clouds is a challenging scenario~\cite{Ma2022ReconstructingSF,du2023rethinking}.
While some of them attempt to utilize extra surface information such as normals~\cite{Qian2021DeepMU,feng2022neural}, they still encounter difficulties in achieving high-fidelity geometry approximation and accurately constructing the surfaces.

\begin{figure*}[t]
    \centering
    \includegraphics[height=5.6cm]{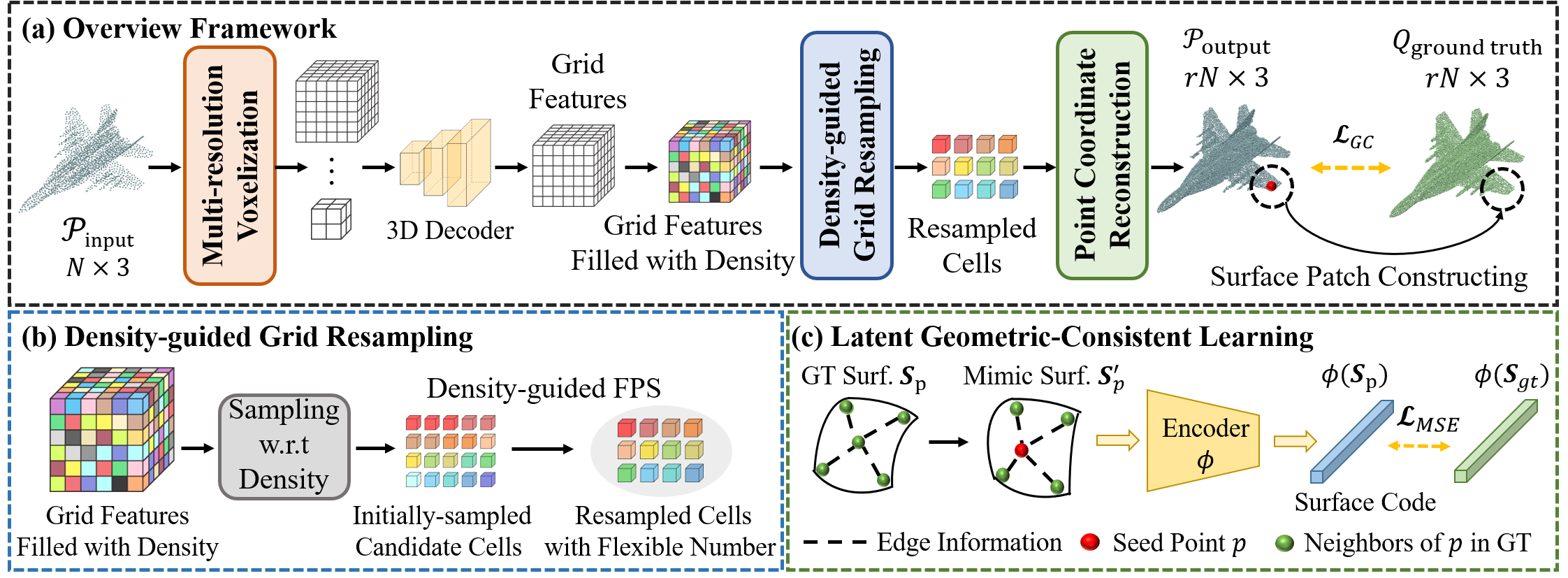}
    \caption{(a) Overview of the proposed PU-VoxelNet. Given an input point set, we first convert it to multi-resolution voxels, and then aggregate the multi-scale voxel representations through a 3D CNN based decoder. Next, the grid cells with the desired output number are sampled by a density-guided resampling strategy. Finally, we reconstruct point coordinates from the sampled grid cells.  (b) Density-guided grid resampling is developed to sample more faithful grid cells while avoiding sampling outliers. (c) Latent geometric-consistent learning focuses on improving the geometry approximation of local surface patches. }
    \label{framework}
\end{figure*}

To address the issues above, we employ voxel grids to model surface representations (Fig.~\ref{figure1}).
Since voxel grids inherently have regular and complete structure~\cite{Mittal2023NeuralRF}, voxel-based networks are capable of providing predefined grid space to represent 3D shapes, and the surface patches within each grid cell can be approximated as a density distribution of points.
So, the learned surface representation is well-constrained within the fixed grids.
Then, we can utilize the density distribution to reconstruct point clouds that closely follow the underlying surface.
To this end, we propose PU-VoxelNet, a novel voxel-based point cloud upsampling network, as illustrated in Fig.~\ref{framework}.
Specifically, we convert point clouds into multiple low-resolution voxels, and aggregate them through a hierarchical 3D convolution neural network.
Then, the surface patches, \textit{i.e.}, filled grid cells, are sampled according to the predicted density distribution, and an arbitrary number of points can be reconstructed from the sampled cells.
However, previous voxel grid sampling methods~\cite{Lim2019ACD,xie2020grnet,wang2021voxel} suffer from inaccurate sampling problem due to the imprecise predictions of density distribution (Fig.~\ref{inaccurate_sampling}).
To address it, leveraging density predictions and grid geometry priors, we present a density-guided grid resampling method to select more faithful samples with fewer outliers.
This approach ensures a high-fidelity point distribution that accurately follows the underlying surface.

In addition, existing training supervisions for upsampling focus on explicitly reducing point-wise errors in the coordinate space~\cite{Yu2018PUNetPC,Li2019PUGANAP}.
Chamfer Distance (CD)~\cite{fan2017point} is a commonly used metric that calculates the average closest point distance between two point sets, which overlooks local point distribution~\cite{wu2021density}.
While some attempts~\cite{Yu2018PUNetPC,Li2019PUGANAP} have been made to fix it by designing auxiliary constraints, they rely on the predefined settings of point distribution and may not generalize well to different distribution patterns.
Here, we aim to learn latent surface geometric consistency between the upsampled and target point clouds.
We observe that the geometry information of the local surface can be effectively embedded into a latent space representation~\cite{Wang2019DynamicGC}.
By measuring the differences in latent surface representations, we can implicitly constrain the points which are not located on the underlying surface.
To this end, using a pretrained surface encoder, we capture the latent geometric discrepancy on a pairs of predefined surface patches.
We formulate the proposed method as an auxiliary training supervision, which can be easily integrated with existing methods.

To demonstrate the effectiveness of our approach, we conduct comprehensive experiments on both synthetic and real-scanned datasets with various settings.
The contributions of this paper are summarized as follows:
\begin{itemize}
    \item We present PU-VoxelNet, a novel point cloud upsampling framework using voxel-based network, to perform upsampling with arbitrary rates.
    \item We investigate the inaccurate sampling problem of previous grid sampling schemes, and propose a density-guided grid resampling approach to generate high-fidelity point distribution while avoiding sampling outliers.
    \item We design a latent geometric-consistent training supervision to constrain local surface patches for further improvements.
    \item The proposed approach is evaluated on various experiments to demonstrate its superiority when compared with the state of the art.
\end{itemize}

\section{Related Work}
\subsection{Learning-based Point Cloud Upsampling}
In this section, we provide a detailed review of learning-based point cloud upsampling methods, focusing on whether they support arbitrary-scale upsampling.

\textbf{Fixed-scale Upsampling.} Earlier learning-based  approaches~\cite{Yu2018PUNetPC,Yu2018ECNetAE,Wang2019PatchBasedP3,Li2019PUGANAP,Qian2020PUGeoNetAG,Qian2021PUGCNPC,Zhao2021SSPUNetSP,Li2021PointCU,yan2022fbnet,Du2022PointCU} are designed for fixed upsampling rates.
As a pioneering work, PU-Net~\cite{Yu2018PUNetPC} proposes a universe upsampling framework, consisting of feature extraction, feature expansion, and point reconstruction.
Subsequent methods follow the same pipeline and employ more advanced point-based networks.
For instance, PU-GCN~\cite{Qian2021PUGCNPC} builds local graphs in the neighborhood of each point to exploit the local structure information.
PU-Transformer~\cite{Qiu2021PUTransformerPC} adopts the general structure of transformer encoder for point feature extraction.
However, one limitation of these methods is that their network architectures are coupled with the upsampling rates. As a result, multiple networks need to be built and trained at different rates. This increases the storage cost and computation burden, making it less practical for real-world applications.

\textbf{Arbitrary-scale Upsampling.} As above-mentioned, the advantages of arbitrary-scale upsampling are highly desirable for various real-world applications~\cite{Ye2021MetaPUAA,dell2022arbitrary}.
To this end, several methods~\cite{Qian2021DeepMU,LuoPUEVAAE} regard upsampling task as a local linear approximation of a curved surface, and generate new points by learning an affine combination of neighbor points.
Certain methods~\cite{feng2022neural,SelfPCU,He_2023_CVPR} focus on learning an implicit surface from point clouds.
Then, they can sample an arbitrary number of points from the surface or push the points to the underlying surface.
Generally, these methods formulate the arbitrary-scale upsampling as a surface approximation problem.
Since the input point clouds are sparse and unorganized, their point-based networks suffer from low fidelity geometry approximation and struggle in accurately building an underlying surface.
In contrast, our method follows the predefined voxel grid space to model the surface as a density distribution of points and thus achieves a better surface approximation for sparse inputs.

\subsection{Voxel Representations in 3D Shape Generation}
Voxel representations can be viewed as a straightforward generalization from the 2D pixel to the 3D domain, which explicitly encode the spatial relationships among points and are suitable for representing the 3D geometry~\cite{mescheder2019occupancy}.
Many approaches~\cite{Wu2016LearningAP,han2017high,Dai2017ShapeCU,groueix2018papier,Lim2019ACD,xie2020grnet,wang2021voxel} have applied this representation for 3D shape generation.
Among them, certain methods~\cite{Lim2019ACD,xie2020grnet,wang2021voxel} take advantages of both point cloud and voxel representations.
They first embed the point cloud into voxel grids, and then apply 3D convolution to learn voxel representations.
Finally, the point cloud is reconstructed from the voxel using the grid sampling strategies which depend on the learnable properties in each grid cell, \textit{e.g.}, vertex combination weights, occupancy classification probability, and point density.
However, due to the unavoidable imprecise predictions, they suffer from inaccurate sampling problem which tends to include outliers, particularly for large upsampling rates.
Therefore, we aim to develop a robust and effective scheme that can mitigate the issue of inaccurate sampling.

\section{The Proposed Approach}
In this section, we first describe the architecture of our voxel-based upsampling network.
Next, we provide details on the proposed density-guided grid sampling method.
Finally, we present a latent geometric-consistent learning and introduce supervisions for end-to-end training.

\subsection{Voxel-based Upsampling Network}
\label{voxel_learning}
Given a sparse point cloud set $\mathcal{P}=\left\{p_{i}\right\}_{i=1}^{N}$ as input, where $N$ is the number of input points, the objective of point cloud upsampling is to generate a dense and high-fidelity point cloud set $\mathcal{Q}=\left\{q_{i}\right\}_{i=1}^{rN}$, where $r$ is the upsampling rate.
In the following, we briefly present each component of our voxel-based point cloud upsampling network. The detailed architecture is given in supplementary materials.

\noindent\textbf{Multi-scale voxelization and aggregation.}
In order to apply the voxel-based network, voxelization is a necessary procedure to regularize unorganized point clouds.
In this work, we follow a similar gridding technique as~\cite{rethage2018fully,Lim2019ACD}.
Firstly, the point displacements $\Delta p_{i}\in\mathbb{R}^{N\times3\times8}$ between each point $p_{i}$ and its nearest eight grid vertexes are computed within four different low-resolution voxels from $4^3$ to $32^3$.
Then, the displacements are encoded into point features using a series of MLPs, and the mean of point features within each grid cell is set as the initial voxel representation.
After that, we aggregate the voxel representation from the started low resolution to the high resolution through a 3D CNN based decoder,
and obtain output voxel representations $\mathcal{F}_\text{g}\in \mathbb{R}^{128\times 32^3}$.

\noindent\textbf{Voxel grid sampling.}
In the following, we approximate the local surface patch as a density distribution of points within each grid cell, and perform grid sampling to collect grid cells that follow the underlying surface.
To this end, a binary classification probability $p_{c}$ and a density value $\delta_{c}$ of each grid cell are learned from voxel representations $\mathcal{F}_\text{g}$.
The probability $p_{c}$ decides whether a grid cell $c$ is filled or empty, and the density $\delta_{c}$ denotes the number of points that should be generated from the non-empty cell.

The grid sampling totally depends on the predictions of $p_{c}$ and $\delta_{c}$, and thus the accuracy of predictions influences the quality of surface reconstruction subsequently.
However, the network inevitably makes imprecise predictions, leading to the inaccurate sampling problem.
This can result in the neglect of desirable grid cells that contain the ground-truth surface patches, while outliers may be unexpectedly selected (Fig.~\ref{inaccurate_sampling_1} and Fig.~\ref{inaccurate_sampling_3}).
Then, it is difficult to further adjust outliers to the underlying surface, thus leading to the low fidelity of point distribution.
As shown in Fig.~\ref{inaccurate_sampling_2}, we can find the sampled cells of previous sampling methods~\cite{Lim2019ACD,wang2021voxel} have large Chamfer Distance to the ground truth, which means they indeed suffer from the inaccurate sampling problem.
To handle this issue, we propose a density-guided grid resampling strategy that takes both density predictions and grid geometry information for sampling.
Specifically, more candidate cells (larger than the desired output number ${rN}$) are first collected by multinomial sampling, and then we design Density-guided Farthest Point Sampling (D-FPS) to reduce the samples to the output number.
As a result, we can obtain a faithful point distribution with fewer outliers.

\noindent\textbf{Point reconstruction.}
Finally, we reconstruct point clouds according to the feature of sampled cells.
To make the sampled features $\mathcal{F}_\text{pt}$ generate different points within a grid cell, the features $\mathcal{F}_\text{pt}$ attached with 2D variables~\cite{groueix2018papier} are firstly used to generate coarse results $\mathcal{P}_\text{c}$ by learning offsets of the corresponding grid cell center.
Then, we adopt a layer of point transformer~\cite{Zhao2020PointT} to predict per-point offsets again, which are used to further adjust the coarse results $\mathcal{P}_\text{c}$.
Thereby, we obtain the final output $\mathcal{P}_\text{r}$ with more local details, and high fidelity.

\subsection{Density-guided Grid Resampling}
In this section, we provide more details and discussions on the proposed density-guided grid resampling method that consists of two sampling stages.
Firstly, multinomial sampling is applied to collect candidate grid cells following the multinomial distribution with respect to filled density $\delta_{c}=\delta_{c} \cdot \text{Sigmoid}(p_{c})$.
So, the probability distribution of sampled points with $n$ times is
\begin{equation}
P\left(X_{j}=n_{j|j=1, \ldots,s}\right)=\frac{n !}{n_{1} ! n_{2} ! \cdots n_{k} !} \delta_{c_1}^{n_{1}} \delta_{c_2}^{n_{2}} \cdots \delta_{c_k}^{n_{k}},
\end{equation}
where $n_{j} \geq 0$ is the times that the $j$th grid cell is sampled, $s$ is the number of all the grid cells, $n$ is the total sampling times and $\sum_{j=1}^{s} n_{j}=n$.
The sampling is drawn in $r'{N}$ independent trials, where $r'$ is the resampling rate which is larger than the target upsampling rate $r$.
Secondly, we need to reduce the samples to the desired output number $r{N}$.
Resorting to the proposed D-FPS, we take both density predictions and grid geometry information for the second sampling.
Specifically, vanilla FPS algorithm maintains a distance array $\mathcal{D}=\left\{d_{i}\right\}_{i=1}^{N}$ that indicates the shortest distance from the $i$-th cell to the already-sampled cell set, and the cell with the largest distance will be sampled  in this time.
Since the density $\delta_{c_i}$ can be regarded as the importance of each cell, to avoid sampling outliers, we incorporate this value into the cell-to-set distance $d_{i}$ as
\begin{equation}
     \hat{d}_{i} = \delta_{c_i}\cdot d_{i},
\end{equation}
where the $\hat{d}_{i}$ is the density-guided cell-to-set distance.

\begin{figure}[t]
\centering
\subfigure[]{\label{inaccurate_sampling_1} \includegraphics[height=3cm]{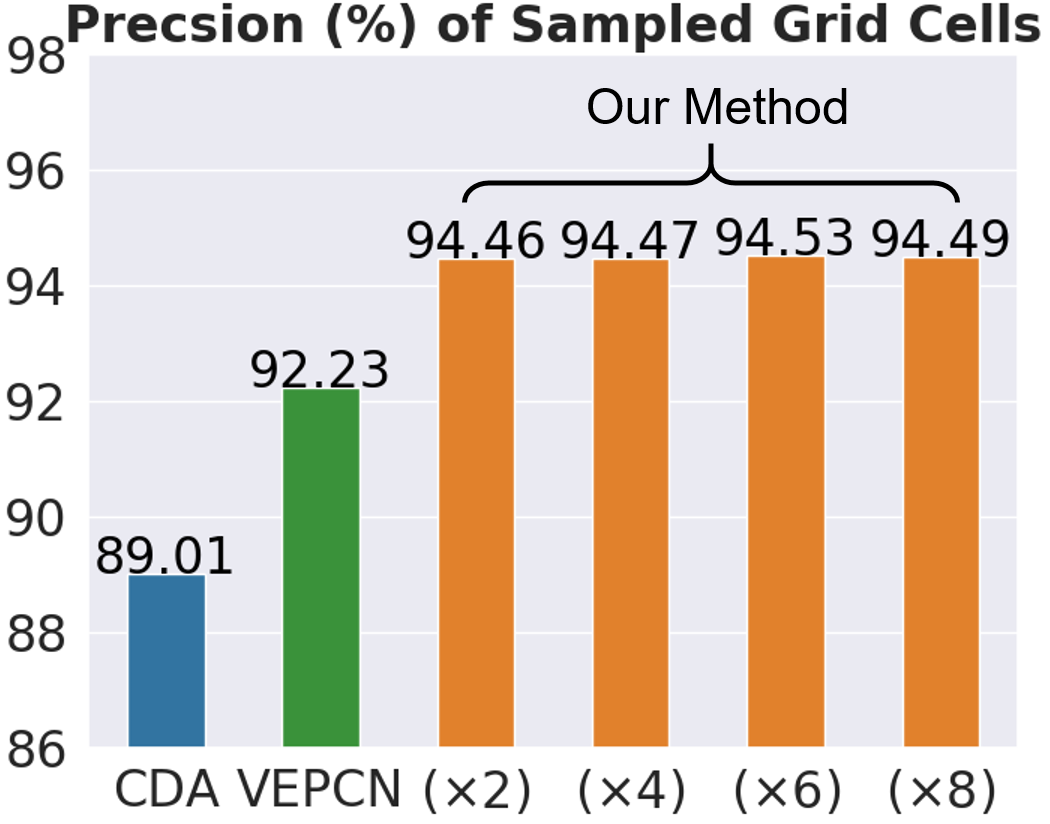}}
\subfigure[]{\label{inaccurate_sampling_2}\includegraphics[height=3cm]{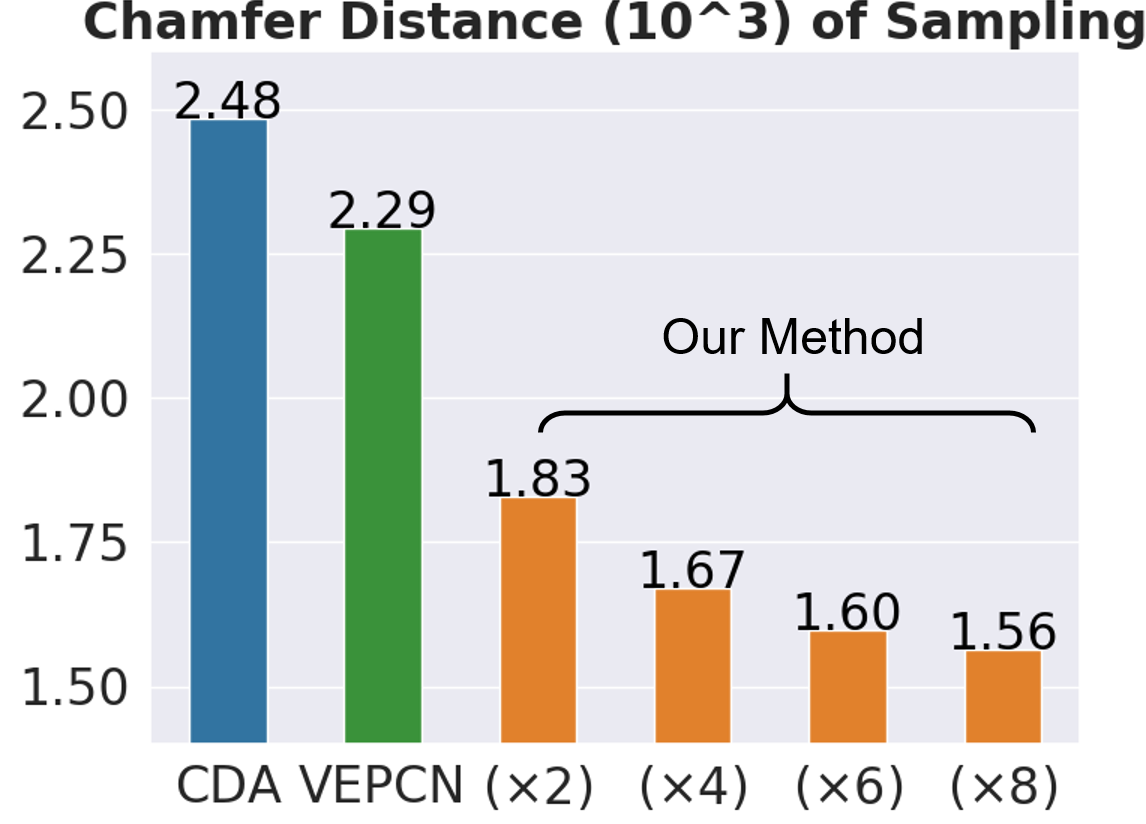}}
\\
\subfigure[]{\label{inaccurate_sampling_3}\includegraphics[height=3cm]{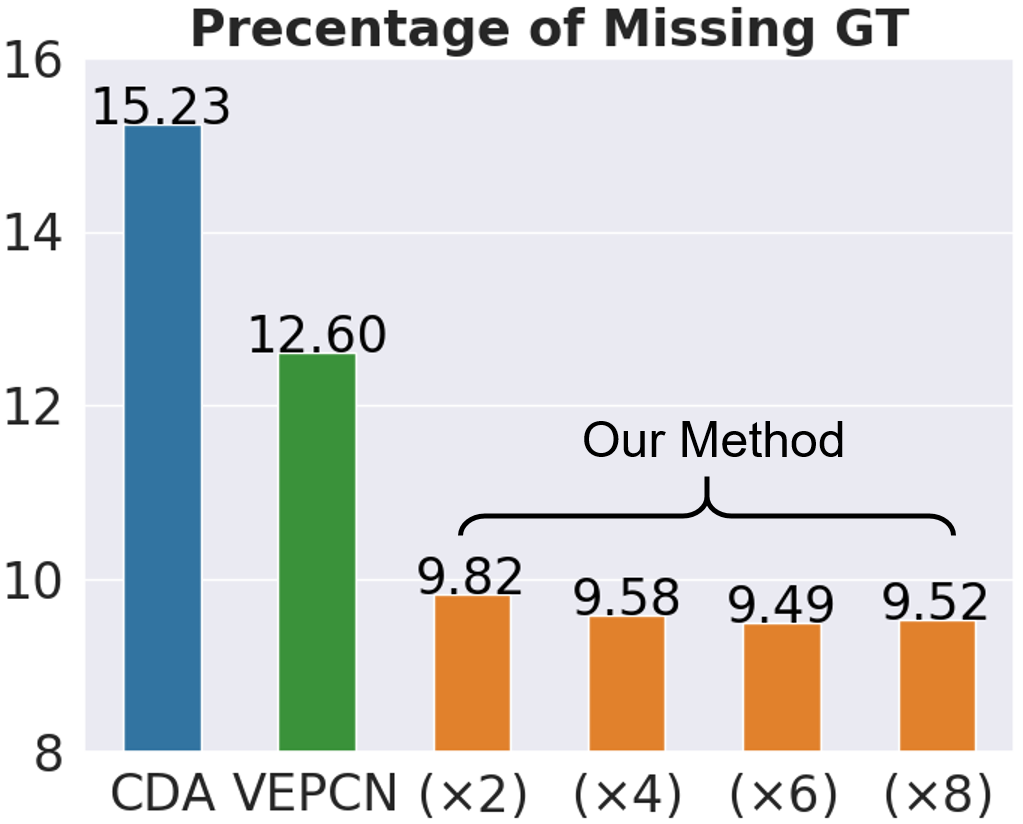}}
\subfigure[]{\label{cd_vs_reasmpling}\includegraphics[height=3cm]{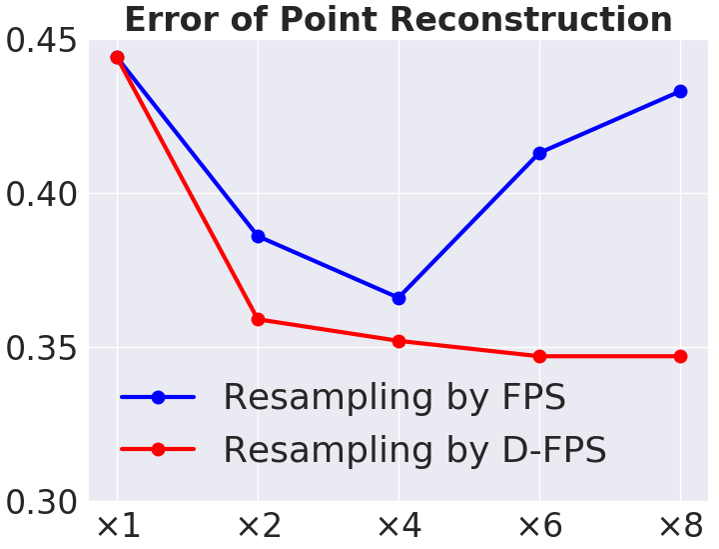}}
\caption{Analysis of the inaccurate sampling problem on PU1K dataset.
Note that we regard the adjacent grid cells as right predictions in this experiment.
(a) Percentage of the ground-truth points in sampled cells.
(b) Chamfer Distance between sampled cells and the ground truth.
(c) Percentage of the ground-truth cells missed by sampling.
Previous methods~\cite{Lim2019ACD,wang2021voxel} (blue and green bar) suffer inferior sampling, \textit{i.e.}, low precision and high missing rate, while our method (orange bars) can sample faithful grid cells which are more close to the underlying surface.
(d) Errors with increasing resampling rates.
The proposed D-FPS effectively avoids sampling outliers, resulting in stable improvements.
}
\label{inaccurate_sampling}
\end{figure}

The advantages of our density-guided grid resampling strategy are two folds.
Firstly, due to the imprecise predictions, previous methods might miss the desirable grid cells and include outliers.
In contrast, the proposed resampling strategy considers both density predictions and fixed grid geometry for sampling, and thus is more robust to the imprecise predictions (Fig.~\ref{inaccurate_sampling}).
Secondly, compared with vanilla FPS, the proposed D-FPS enables to preserve faithful grid cells and be more robust to outliers.
While the outliers have a large distance to other points, they usually have a small density, which will not be chosen by D-FPS.
The green line in Fig.~\ref{cd_vs_reasmpling} validates the effectiveness of D-FPS.
Benefiting from the advantages above, our sampling strategy shows better capacity of point cloud upsampling, especially on large upsampling rates (as shown in the ablation study).

\subsection{Latent Geometric-Consistent Learning}
Most previous approaches employ Chamfer Distance (CD) or Earth Mover’s Distance (EMD)~\cite{fan2017point} to reduce the point-to-point distance errors.
Their drawbacks have been mentioned by prior works~\cite{xie2020grnet,wu2021density}, such as insensitivity to the local point distribution.
Although some attempts~\cite{Yu2018PUNetPC,Li2019PUGANAP} have been made to design auxiliary constraints, they rely on the prior settings of point distribution, and thus cannot generalize well to different distribution patterns.

In this work, we consider the constraint on latent representations is also helpful to improve fine-structured locality, since the local neighborhood structure, \textit{i.e.}, edge information, can be well embedded into latent space~\cite{Wang2019DynamicGC}.
By measuring the differences of the latent surface representations, we can learn geometric consistency between two surface patches.
To fulfill this objective, the remained problem is how to construct an effective surface representation, since it is difficult to directly construct surface from raw points without extra information.

Here, we regard the ground-truth points as subset of the underlying surface. Then, we present an approximated way that regards each upsampled point ${p}$ as a seed point and then finds its neighbor points ${N}({p}, {Q})$ from the ground truth ${Q}$.
The $k$-nearest neighbors are grouped as the real (ground-truth) surface patch ${S}({p},{Q} )=\left\{s_{i}\right\}_{i=1}^{k} , s_{i} \in {N}({p}, {Q})$, and $s_{1}$ is the nearest point to ${p}$.
Then, we replace $s_{1}$ in ${S}({p},{Q} )$ with its seed point ${p}$ to obtain a mimic surface patch ${S}'({p}, {Q})$.
Such point replacement scheme will cause the change of latent surface representations, if the seed point is not located in the right position, such as out of the underling surface.
Then, we construct a mapping between Cartesian coordinates $ \mathbb{R}^{3\times N}$ and latent space $ \mathbb{R}^{D}$,
\begin{equation}
\phi(\cdot): \mathbb{R}^{3\times N} \rightarrow \mathbb{R}^{D}.
\end{equation}
In practical implementation, a pre-trained surface encoder is adopted as $\phi(\cdot)$ to capture the feature-wise differences between mimic and real local surface patches. Hence, we define the geometric-consistent learning loss as:
\begin{equation}
  \mathcal{L}_\text{GC}\left({P}, {Q}\right)
  = \frac{1}{|{P}|} \sum_{p \in {P}} || \phi({S}({p},{Q})) - \phi({S}'({p},{Q}))||,
\end{equation}
where ${P}$ and ${Q}$ denote the generated point clouds and their corresponding ground truth, respectively. By doing so, we can capture the latent geometric discrepancy to push the seed points to lie on the target surface.
The detailed implementation can be found in our supplementary materials.

\begin{figure}[t]
    \centering
    \includegraphics[height=2.8cm]{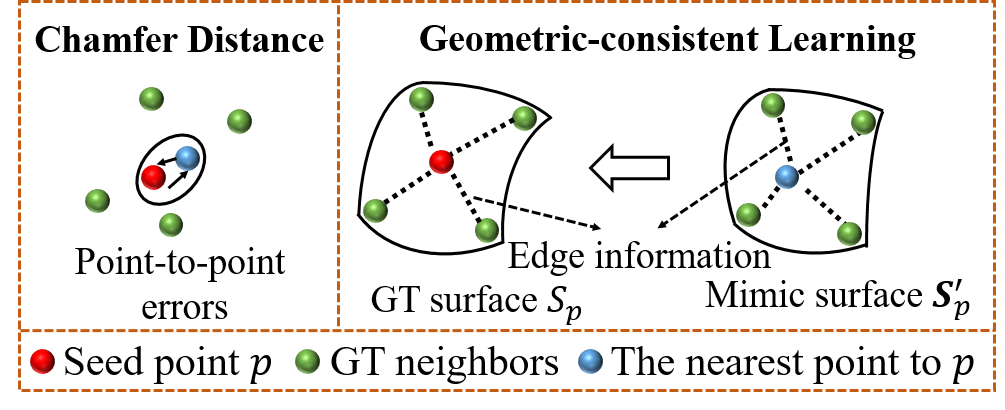}
    \caption{A toy example shows the differences between Chamfer Distance (CD) and our method. CD only penalizes on point-wise errors, which fails to exploit the relationships within neighbor points, and cannot guarantee the local structure well. In contrast, we constrain the ``edge'' information around a seed point, which is complementary to improve  fine-structured locality of surface patches. }
    \label{geo_loss}
\end{figure}

\subsection{End-to-end Training}
Our PU-VoxelNet is trained using the proposed geometric-consistent loss, the point-wise loss and the voxel-based loss.

\textbf{Point-wise loss.}
To encourage the upsampled points to distribute over the target surface, we adopt CD loss $\mathcal{L}_\text{CD}$ and its sharp version $\mathcal{L}^\text{S}_\text{CD}$~\cite{Lim2019ACD}:
\begin{equation}
\small
  \mathcal{L}_\text{CD}\left(\mathcal{P}, \mathcal{Q}\right)
  = \frac{1}{|\mathcal{P}|} \sum_{p \in \mathcal{P}} \min _{q \in \mathcal{Q}} \left\|p-q\right\|^{2}_{2} + \frac{1}{|\mathcal{Q}|} \sum_{q \in \mathcal{Q}} \min _{p \in \mathcal{P}} \left\|p-q\right\|^{2}_{2},
\end{equation}
We adopt CD loss for both coarse and refined results, and its sharp version only for coarse results. Since we aim to predict the per-point offset on the corresponding cell center $c_\text{o}$, the generated points should be distributed within center's neighbours. Then, as did in~\cite{Lim2019ACD}, a regularization loss is used to penalty the outlier points, which is defined as $\mathcal{L}_\text{reg}(\mathcal{P})=\sum_{c} \sum_{p \in \mathcal{P}} \max \left(\operatorname{dist}\left(p, c_{o}\right)-d, 0\right)$, and $d$ is the diagonal length of grid cells.

\textbf{Voxel-based loss.} CD loss only penalizes the point-wise differences. To promote the approximated surface patches follow the ground truth, we adopt a binary cross entropy loss $\mathcal{L}_\text{BCE}(\cdot)$ on probability $p_{c}$ with its ground truth $\hat{p}_{c}$, and a mean squared error loss  $\mathcal{L}_\text{MSE}(\cdot)$ between predicted density $\delta_{c}$ and its ground truth $\hat{\delta}_{c}$.
Thus, the total loss function is
\begin{equation}
\begin{aligned}
\small
 \mathcal{L}_\text{total} & =
\mathcal{L}_\text{CD}\left(\mathcal{P}_\text{c}, \mathcal{Q}\right) +
\lambda_{1} \mathcal{L}^\text{S}_\text{CD}\left(\mathcal{P}_\text{c}, Q\right) + \mathcal{L}_\text{CD}\left(\mathcal{P}_\text{r}, \mathcal{Q}\right)
\\& +
 \lambda_{2} \mathcal{L}_\text{GC}\left(\mathcal{P}_\text{r}, \mathcal{Q}\right) + \lambda_{3}\mathcal{L}_\text{reg}(\mathcal{P}_{c}) + \lambda_{4}\mathcal{L}_\text{BCE}(p_c, \hat{p}_c)
\\& + \lambda_{5}
\mathcal{L}_\text{MSE}(\delta_c, {\hat\delta}_c),
\end{aligned}
\end{equation}
where $\lambda_{1},\lambda_{2},\lambda_{3},\lambda_{4} ,\lambda_{5} $ are balance weights.

\section{Experiments}

\subsection{Datasets and Experimental Settings}
\textbf{Datasets.}
To make the experiments reproducible, we utilize two public datasets with their settings directly, including PU-GAN~\cite{Li2019PUGANAP} and PU1K~\cite{Qian2021PUGCNPC}.
In addition, we also employ a real-scanned dataset,~\textit{i.e.}, ScanObjectNN~\cite{Uy2019RevisitingPC}, for qualitative evaluation.

\textbf{Training details.}
Our models are trained by 100 epochs with a batch size of 64 on PU1K dataset, and a batch size of 32 on PU-GAN dataset.
The learning rate begins at 0.001 and drops by a decay rate of 0.7 every 50k iterations.
As did in ~\cite{Li2019PUGANAP,Qian2021PUGCNPC}, the  training patch contains 256 points, and the corresponding ground truth contains 1,024 points.
For loss balanced weights, we empirically set $\lambda_{1} = 300$, $\lambda_{2} = 0.01$, $\lambda_{3} = 0.3$, $\lambda_{4} = 100$, $\lambda_{5} = 1e^{10}$. The resampling rate is 4, and $k$ is 16 in surface patches.

\textbf{Evaluation.}
For a fair evaluation, we follow the same configurations as PU-GCN~\cite{Qian2021PUGCNPC} and Grad-PU~\cite{He_2023_CVPR}.
We first select the seed points from the input test point clouds by FPS and group the multiple local patches using $K$-NN based on the seed points.
Then, these local patches are upsampled with $r$ times.
Finally, all the overlapping patches are merged and down-sampled as the outputs by FPS.
Three commonly-used evaluation metrics are adopted for quantitative comparison, including Chamfer Distance (CD), Hausdorff Distance (HD), and Point-to-Surface Distance (P2F).

\textbf{Comparison methods.} We make comparison with five learning-based methods designed for fixed upsampling rates, including PU-Net~\cite{Yu2018PUNetPC}, MPU~\cite{Wang2019PatchBasedP3}, PU-GAN~\cite{Li2019PUGANAP}, Dis-PU~\cite{Li2021PointCU}, PU-GCN~\cite{Qian2021PUGCNPC}. Furthermore, we compare with MAFU~\cite{Qian2021DeepMU}, PU-SSAS~\cite{SelfPCU} and Grad-PU~\cite{He_2023_CVPR}, which  also support flexible rates.

\begin{figure*}[t]
    \centering
    \includegraphics[scale=0.215]{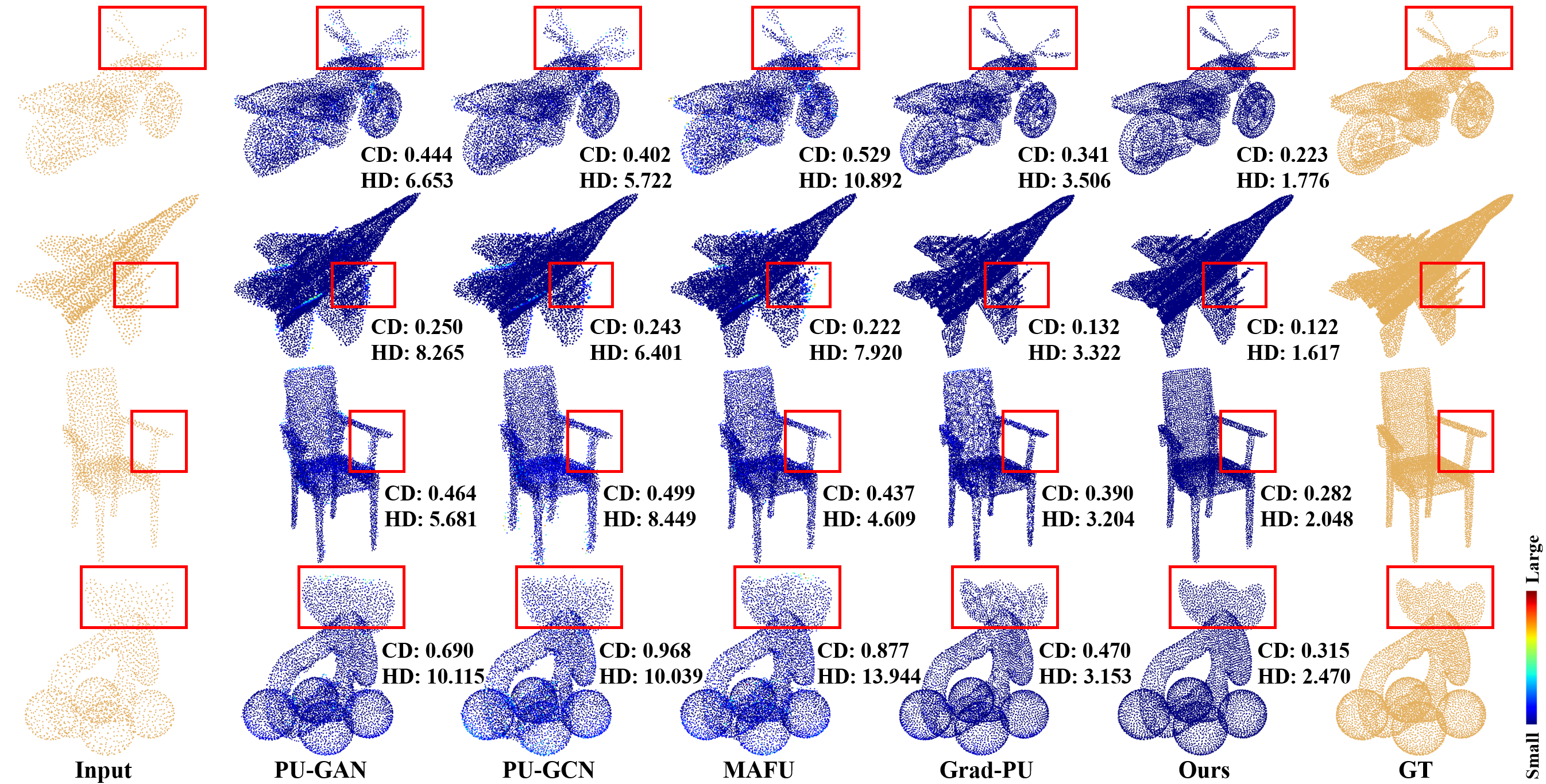}
    \caption{Upsampling ($4\times$) results on PU1K dataset with input size of 2,048. The points are colored by the nearest distance between the ground truth and the upsampled points. The blue denotes the small errors. One can zoom in the figure for details.}
    \label{puk1_vis}
\end{figure*}

\begin{table}[t]
    \begin{center}
    \resizebox{0.9\linewidth}{!}{
    \begin{tabular}{l|p{1cm}<{\centering}p{1cm}<{\centering}p{1cm}<{\centering}}
    \toprule[1pt]
  {Methods}&  {CD}& {HD}&  {P2F}\\
    \midrule[0.5pt]
    PU-Net~\cite{Yu2018PUNetPC}  &1.157 & 15.297 & 4.924 \\
    MPU~\cite{Wang2019PatchBasedP3}  &0.861 & 11.799 &3.181 \\
    PU-GAN~\cite{Li2019PUGANAP}  &0.661 &9.238 & 2.892 \\
    Dis-PU~\cite{Li2021PointCU}&0.731&9.505&2.719  \\
    PU-GCN~\cite{Qian2021PUGCNPC} &0.585& 7.577 &2.499 \\
    MAFU~\cite{Qian2021DeepMU} &0.670&10.814&  2.633\\
    Grad-PU~\cite{He_2023_CVPR}&\underline{0.404} &\underline{ 3.732} & \underline{1.474}\\
  \midrule[0.5pt]
    PU-VoxelNet (Ours) & \textbf{0.338} &\textbf{2.694}&\textbf{1.183} \\
    \bottomrule[1pt]
    \end{tabular}}
    \caption{Quantitative comparison ($4\times$ upsampling) on PU1K dataset with different input sizes of point clouds. The values of CD, HD, and P2F are multiplied by $10^{3}$.  }
    \label{PU1K_performance}
    \end{center}
\end{table}

\subsection{Comparison on Synthetic Dataset}
In this section, we aim to demonstrate the superiority of PU-VoxelNet over SOTA methods on two synthetic datasets.

\textbf{Results on PU1K dataset.}
Table~\ref{PU1K_performance} reports quantitative comparisons on PU1K datasets.
Overall, we consistently achieve evident improvements over other counterparts, indicating the advantage of our approach.
Benefiting from the predefined spaces of regular grids, the upsampling becomes controllable to approximate object shapes by geometry constrains, and thus generates high-fidelity point clouds.
Fig.~\ref{puk1_vis} gives some visualization results. We can clearly notice existing methods have outliers and lack of fine-grained details on complex geometry structure.
For comparison, PU-VoxelNet is able to generate uniform point clouds with better details.

\begin{table}[t]
\begin{center}
    \resizebox{0.9\linewidth}{!}{
    \begin{tabular}{l|ccc|ccc}
    \toprule[1pt]
  \multirow{2}{*}{Methods}
  & \multicolumn{3}{c|}{{$4\times$ Upsampling} }
  & \multicolumn{3}{c}{{$16\times$ Upsampling}}  \\
  & {CD}& { HD} & {P2F}& {CD} & { HD} & {P2F} \\
    \midrule[0.5pt]
     PU-Net&0.401 &4.927&4.231 &0.323& 5.978& 5.445 \\
    MPU &0.327 &4.859 &  3.070&0.194&6.104 &3.375 \\
    PU-GAN & 0.281&4.603&3.176&0.172 &5.237 &3.217  \\
    Dis-PU & 0.265& 3.125& 2.369&0.150 &3.956 &2.512 \\
    PU-GCN & 0.268&3.201& 2.489&0.161 & 4.283& 2.632 \\
    MAFU & 0.285&3.976&2.314&0.156 &4.288 &2.644\\
    PU-SSAS & 0.251&3.491& 2.163&0.153&3.696 &2.513 \\
    Grad-PU & \underline{0.245}&\underline{2.369} &\textbf{1.893}&\underline{0.108}&\underline{2.352}  &\textbf{2.127}\\
    \midrule[0.5pt]
    Ours &\textbf{0.233}&\textbf{1.751}& \underline{2.137}&\textbf{0.091}&\textbf{1.726}& \underline{2.301}\\
    \bottomrule[1pt]
    \end{tabular}}
    \caption{Quantitative comparison on PU-GAN dataset with two different upsampling rates. The values of CD, HD, and P2F are multiplied by $10^{3}$.
    }
    \label{PU_GAN_performance}
    \end{center}
\end{table}

\begin{figure}[t]
\subfigure[Chamfer Distance ($10^{3}$)]{\label{flexible_rate_cd} \includegraphics[height=3.2cm]{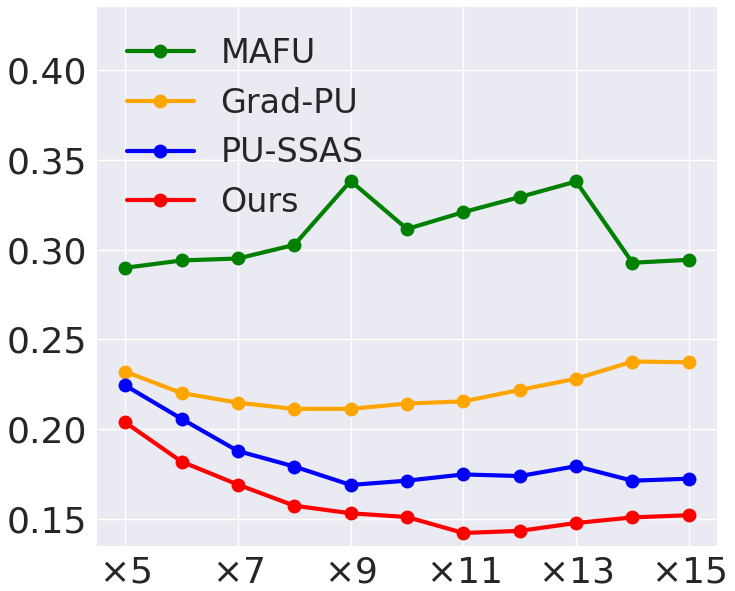}}
\subfigure[Hausdorff  Distance ($10^{3}$)]{\label{flexible_rate_hd}\includegraphics[height=3.3cm]{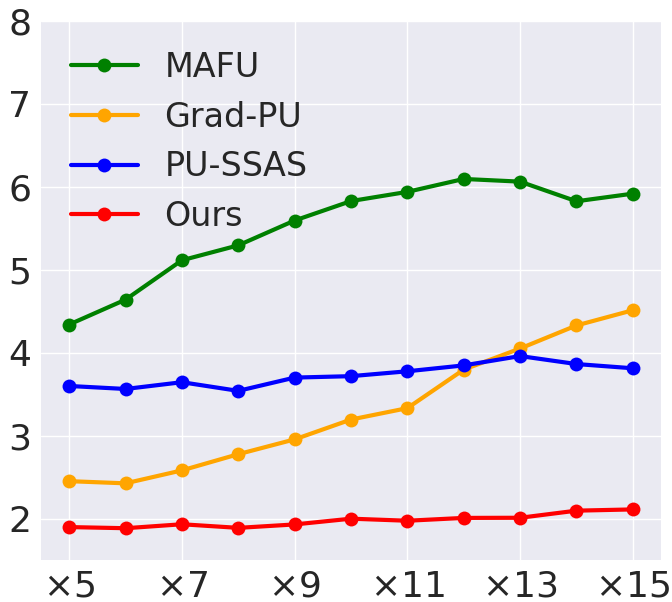}}
\caption{ Arbitrary-scale upsampling on PU-GAN dataset. }
\label{flexible_up}
\end{figure}

\textbf{Results on PU-GAN dataset.}
Moreover, we conduct experiments with two upsampling rates ($\times$4 and $\times$16) on PU-GAN dataset~\cite{Li2019PUGANAP}.
Since PU-GAN dataset only provides training patches based on $\times$4 upsampling rate, we apply all the model twice for $\times$16 upsampling in this experiment.
As shown in Table~\ref{PU_GAN_performance}, our method outperforms other counterparts in most cases, implying that PU-VoxelNet has better generalization ability on different upsampling rates.

\subsection{Arbitrary-scale Upsampling}
In this section, we make a comparison with MAFU~\cite{Qian2021DeepMU}, PU-SSAS~\cite{SelfPCU} and Grad-PU~\cite{He_2023_CVPR} for arbitrary-scale upsampling.
Here, we only change the upsampling rate and fix other parameters.
From Fig.~\ref{flexible_up}, we can observe the overall errors of PU-VoxelNet (the red line) are reduced as the
increasing upsampling rate, while other methods might fail to accurately construct the underlying surface and thus their Hausdorff Distances degenerate at large upsampling rates.
Our performance is consistently better than other counterparts as the increasing rates, implying the superiority of our voxel-based surface approximation for arbitrary-scale point cloud upsampling.

\subsection{Impact on Surface Reconstruction}
To verify the impact on the downstream task, \textit{i.e.}, surface reconstruction, we apply BallPivoting~\cite{bernardini1999ball} to reconstruct meshes from the points. Then, we follow a common way~\cite{ma2021neural} that computes CD metric between the ground-truth and reconstructed meshes. Table~\ref{reconstruction} reports quantitative comparison on PU-GAN dataset.
The results show that our method obtains a better reconstructed surface than other counterparts and is comparable to the high-resolution result sampled from the ground-truth mesh.

\begin{table}[t]
    \centering
    \resizebox{0.95\linewidth}{!}{
    \begin{tabular}{lccccc}
    \toprule[1pt]
  \multirow{4}{*}{\makecell[c]{Methods \\ CD ($10^{3}$)}}
  &\makecell[c]{Low-res\\ 0.695}
  &\makecell[c]{High-res\\ 0.203} & \makecell[c]{MPU\\ 0.643}  & \makecell[c]{PU-GAN\\ 0.417} &\makecell[c]{Dis-PU \\ 0.397} \\
   \cmidrule{2-6}
 & \makecell[c]{PU-GCN \\ 0.448} & \makecell[c]{MAFU \\ 0.517} & \makecell[c]{PU-SSAS\\ 0.599} & \makecell[c]{Grad-PU \\ 0.376} & \makecell[c]{Ours \\ \textbf{0.245}} \\
    \bottomrule[1pt]
    \end{tabular}}
    \caption{Surface reconstruction on PU-GAN dataset. Low-res (2,048) and High-res (8,192) represent the points downsampled from the ground-truth meshes with two resolutions.
    }
    \label{reconstruction}
\end{table}

\begin{figure}[t]
    \centering
    \includegraphics[scale=0.15]{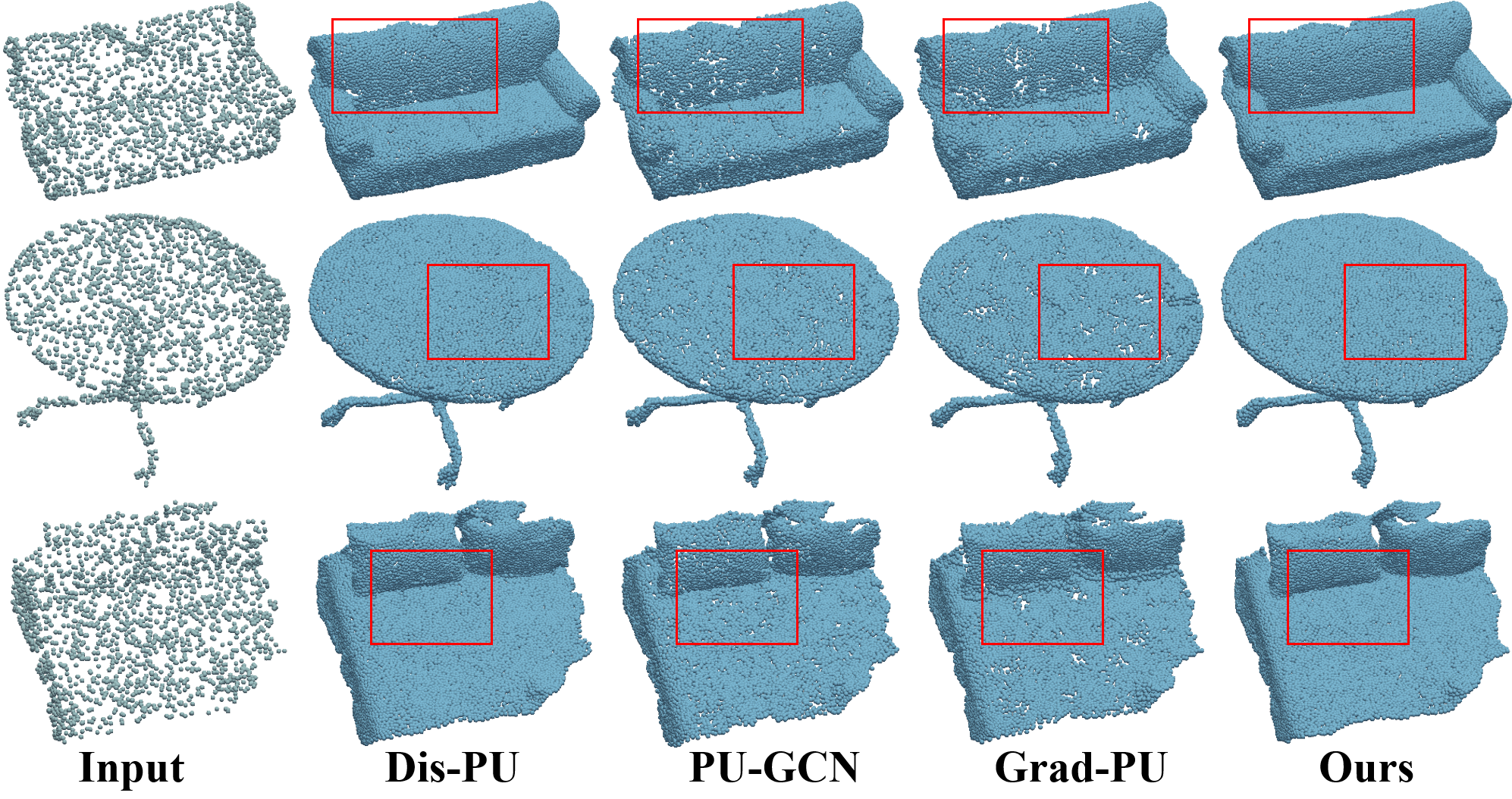}
    \caption{Upsampling ($4\times$) results on real-scanned inputs. }
    \label{vis_3}
\end{figure}

\begin{table}[t]
\centering
\resizebox{0.975\linewidth}{!}{
\begin{tabular}{c|ccc|ccc}
\toprule[1pt]
\multirow{2}{*}{\makecell[c]{Sampling \\ Method}}

  &\multicolumn{3}{c|}{$4\times$ Upsampling}
  &\multicolumn{3}{c}{{$16\times$ Upsampling}}  \\
 &{CD }&{HD}&{P2F}& {CD }&{HD}&{P2F}\\
\midrule[0.5pt]
Original &0.444&3.966&1.533&0.532 &6.166&1.904\\
Ours (FPS)&0.366 &3.123&1.325&0.433 & 4.669&1.396\\
\midrule[0.5pt]
Ours (D-FPS)&\textbf{0.352} &\textbf{2.809} &\textbf{1.283}& \textbf{0.306} &\textbf{3.567} &\textbf{1.307}\\
\bottomrule[1pt]
\end{tabular}}
\caption{Ablation study on density-guided grid resampling.}
\label{ablation}
\end{table}

\begin{table}[ht!]
\centering
    \resizebox{0.975\linewidth}{!}{
    \begin{tabular}{c|ccc|ccc}
    \toprule[1pt]
      \multirow{2}{*}{\makecell[c]{Surface \\  Constructor}}
      &\multicolumn{3}{c|}{Poisson Sampling}
      &\multicolumn{3}{c}{Random Sampling}  \\
     &{CD }&{HD}&{P2F}& {CD }&{HD}&{P2F}\\
    \midrule[0.5pt]
    w/o GC& 0.352& 2.809&1.283 & 0.506 &5.893 & 1.821 \\
    SurRep&0.350 &\textbf{2.646}&1.224& 0.498& 5.665&1.698\\
    \midrule[0.5pt]
    Ours &\textbf{0.338} &2.694 &\textbf{1.183}&\textbf{0.475} &\textbf{5.465} &\textbf{1.646} \\
    \bottomrule[1pt]
    \end{tabular}}
    \caption{Ablation of latent geometric-consistent learning.
    }
    \label{GC_performance}
\end{table}

\subsection{Comparison on Real-scanned Dataset}

To verify the effectiveness of our method in real-world scenarios, we utilize the models trained on PU1K dataset to conduct qualitative experiments on ScanObjectNN~\cite{Uy2019RevisitingPC}.
From Fig.~\ref{vis_3}, we can observe upsampling real-scanned data is more challenging, since the input points are noisy and non-uniform with incomplete regions.
Nevertheless, compared with other competitors, PU-VoxelNet generates more uniform point clouds with more fine details, implying that our model generalizes well on real-scanned data.

\subsection{Ablation Study}
\label{sec_ablation}
In this section, we conduct ablation studies on PU1K dataset.

\textbf{Density-guided grid resampling.}  We first conduct ablation studies on the proposed density-guided grid resampling.
In Table~\ref{ablation}, the first line is original method that only uses multinomial sampling according to the density distribution, and the second line denotes using vanilla FPS for resampling.
The proposed resampling (D-FPS) enables to obtain significant improvements, especially on the large upsampling rate ($16\times$).
The results imply that the proposed density-guided grid resampling greatly mitigates the inaccurate sampling issue when there are a large number of points.

\textbf{Latent geometric-consistent learning.}
To verify the effectiveness of latent geometric-consistent learning, we further conduce experiments under two different input sampling strategies, including Poisson sampling and random sampling.
Here, we also try another way of surface constructor, \textit{i.e.}, ``SurRep'', which denotes minimizing the surface representation~\cite{ran2022surface} between upsampled and ground-truth patches directly.
The results in Table~\ref{GC_performance} show that we can further improve upsampling by constraining surface patches in latent space, and the performance benefits more from the point replacement scheme.
Moreover, we apply the proposed method on different point cloud upsampling methods to explore its wide applicability. Please refer to the supplements for more results.

\section{Conclusion}
In this study, we present PU-VoxelNet, a voxel-based network to accomplish point cloud upsampling with arbitrary rates.
Benefiting from the 3D grid space, the proposed method approximates the surface patches as a density distribution of points within each cell.
To address the inaccurate sampling problem, we develop a density-guided grid resampling method to collect more faithful points with fewer outliers, and thus greatly improves the fidelity of upsampled point distribution.
Furthermore, we propose a latent geometric-consistent learning approach to improve the local geometry approximation of surface patches.
Comprehensive experiments on various settings demonstrate the superiority of PU-VoxelNet over the state-of-the-art methods.

{
  \bibliographystyle{aaai24}

}

\newpage
\appendix

In the supplementary materials, we provide more implementation details, including the network architecture and the latent geometric-consistent learning.
For a through evaluation, we conduct more experiments, including ablation studies on training loss functions, noise robustness test, and more visualization results.

\section{Implementation Details}
In this section, we present details of the network architecture and  geometric-consistent surface learning.

\subsection{Architecture of 3D CNN Decoder}
\begin{figure*}[ht]
    \centering
    \includegraphics[scale=0.4]{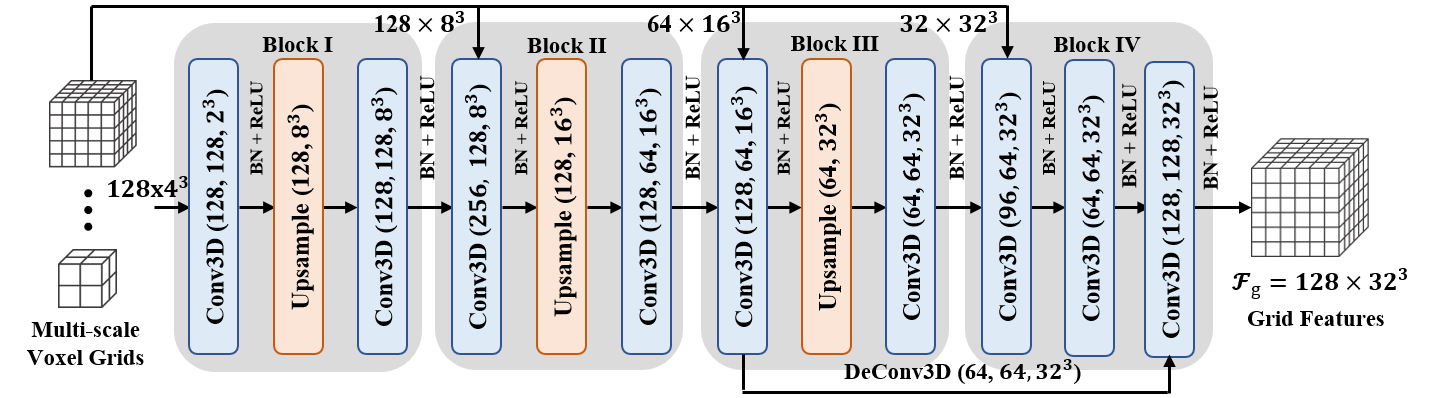}
    \caption{The detailed network architecture of our 3D convolutional decoder for multi-scale voxel aggregation. Firstly, the point displacements between each point and its nearest eight grid vertexes are computed within four resolution voxels ranging from $4^3$ to $32^3$. These displacements are further encoded into point features using three layers of MLPs, and the mean features are set as the voxel representations corresponding to each resolution.
    Then, we aggregate the voxel representation from low resolution to high resolution.
    Finally, we obtain the output voxel representations $\mathcal{F}_\text{g}\in \mathbb{R}^{128\times 32^3}$.  }
    \label{3Ddecoder}
\end{figure*}

The network architecture of our 3D CNN decoder is shown in Fig.~\ref{3Ddecoder}.
The three four blocks contain two 3D convolutional layers with a upsampling layer.
In particular, Conv3D ($128,128,4^3$) denotes a plain 3D convolutional layer with the number of input channel as 128, the number of output channel as 128, and the output voxel resolution as $4^3$.
For all Conv3D layers, we set the kernel size as 3, the stride and padding values as 1.
Besides, Upsample indicates an upsampling operation with the scale factor as 2.
At the beginning of middle blocks (II, III and IV), we concatenate the output features from the previous block and the voxel features from the original multi-scale voxelization module.
Finally, a 3D transposed convolutional (DeConv3D) branch is further used to add variations on the upsampled features at the last block.
For DeConv3D layer, we set the kernel size as 4, the stride as 2, and the padding as 1.

\subsection{Architecture of Point Reconstruction}
To reconstruct point clouds from the sampled gird features, we first add a 2D variables~\cite{groueix2018papier} on these features to promote them generate different points within a grid cell.
By doing so, we obtain the point-wise features $\mathcal{F}_\text{pt}=130 \times rN$.
First, we regress the point offsets using a series of Multilayer Perceptrons (MLPs), with output channel numbers of 128, 64, 32, and 3.
The predicted point offsets are then added on the corresponding cell centers, resulting in the coarse upsampling results $\mathcal{P}_\text{c}$.
Subsequently, the point features $\mathcal{F}_\text{pt}$ are reused to refine the coarse results through a single point transformer layer~\cite{Zhao2020PointT}.
The channel number of hidden layer is 64 in the Transformer.
Finally, we further learn the per-point offset based on the coarse points to obtain refined results  $\mathcal{P}_\text{r}$ by a chain of MLPs, with output channel numbers of 128, 64, 32, and 3.

\begin{figure}[t]
\subfigure[]{\label{overall_encoder} \includegraphics[height=1.9cm]{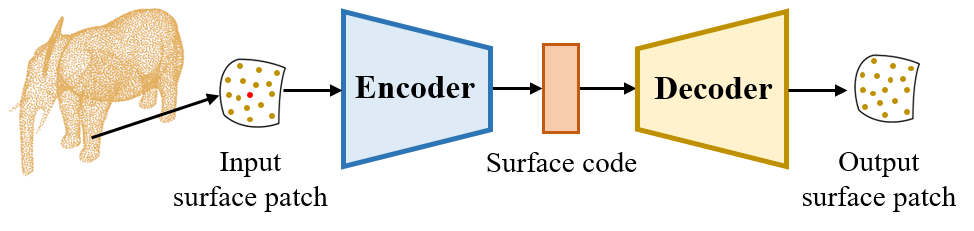}}
\hfill
\centering
\subfigure[]{\label{detailed_encoder}\includegraphics[height=4.5cm]{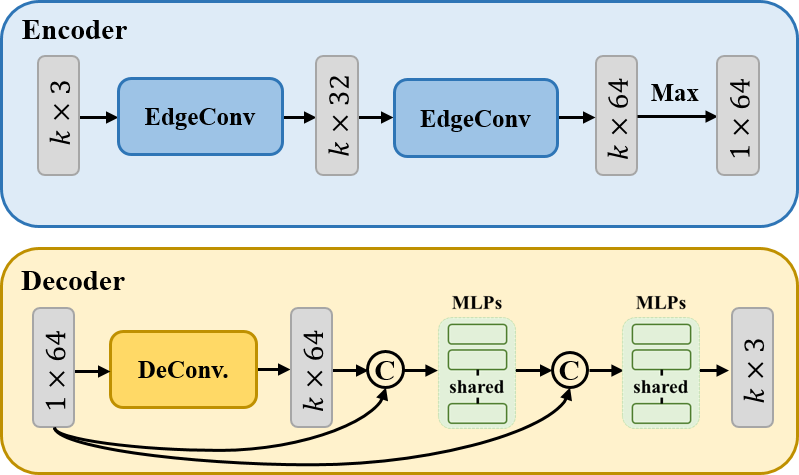}}
\caption{(a) An illustration of surface auto-encoder. The encoder learns to embed the input surface patch into a code, and the decoder aims at reconstructing the surface from the code.
(b) Detailed network architecture of surface encoder and decoder. After the training, we only employ the encoder for latent geometric-consistent surface learning.  }
\label{surface_encoder}
\end{figure}

\begin{figure}[t]
\subfigure[Geometric Consistency]{\label{pt_cd} \includegraphics[height=3.35cm]{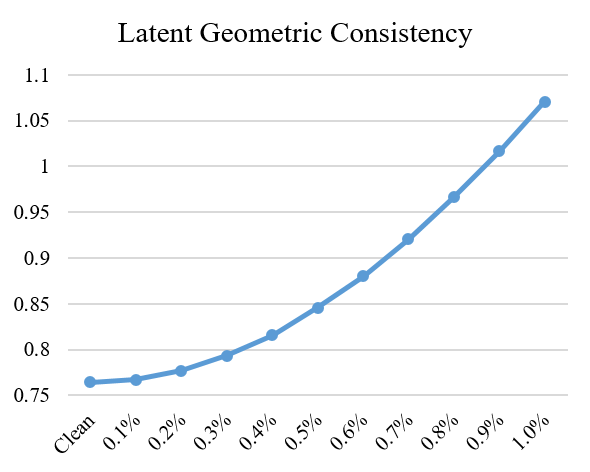}}
  \hspace{-0.5em}
\subfigure[Chamfer Distance]{\label{pt_gc}\includegraphics[height=3.35cm]{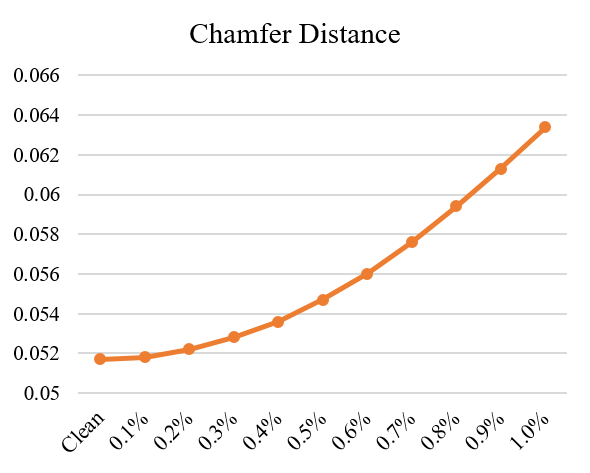}}
 \vskip -10pt
\caption{Chamfer Distance and latent geometric consistency with increasing levels of point perturbation (Gaussian noise on the seed point). The values of LGC and CD are multiplied by $10^{3}$.
  }
\label{surface_loss}
\end{figure}

\subsection{Latent Geometric-consistent Learning}
To embed the geometric information of point clouds, we adopt a pre-trained surface encoder that serve as a mapping between Cartesian coordinate and latent space. Here, we provide additional implementation details about the surface encoder.
Fig.~\ref{surface_encoder} illustrates the pipeline of surface auto-encoder and the detailed network architecture.
Specifically, the encoder consists of two EdgeConv~\cite{wang2019dynamic} layers that aggregate neighboring features $\mathcal{F}_p \in \mathbb{R}^{N \times D} $. Then, we reduce them into a global surface code $\mathcal{I} \in \mathbb{R}^{1 \times D} $ by max pooling.
As did in ~\cite{xiang2021snowflakenet}, the decoder applies a transposed convolutional layer to recover the point features that are further integrated with duplicated surface code for point reconstruction.

Thanks to the EdgeConv-based surface encoder, we are able to construct latent surface representations that embed the ``edge' information of the local neighborhood patches. This allows the encoded surface representations to effectively capture and distinguish subtle point perturbations.
As shown in Fig.~\ref{surface_loss}, it can be observed that the geometric consistency between the mimic surface and the ground-truth surface is enlarged as the level of point perturbation increases. Additionally, we can notice that there is a positive relationship between geometric consistency and Chamfer Distance. This implies that penalizing this metric can also help in reducing Chamfer Distance errors.

To achieve this, the learning objective of surface encoder is to embed the geometric information of local surface patches, which is represented by a surface code.
During the training, we sample $M$ surface centers $c_i$ from the input point cloud using FPS algorithm, and then build a local surface patch around each center with $K$-NN search.
Finally, each training local surface patch contains $k$ points.
$M$ and $k$ are set as 64 and 16 in our experiment, respectively.
We employ CD loss as training loss function for point cloud reconstruction.
The training data is collected from PU1K dataset~\cite{Qian2021PUGCNPC}.
The entire model is trained for 100 epochs with a batch size of 512. The learning rate starts at 0.001 and is decayed by 0.7 every 35k iterations.
After the training, we only utilize the surface encoder for latent geometric-consistent learning.

\section{More Experimental Results}
In this section, we provide more experimental results, including more ablation studies on training loss functions, noise robustness test, and more visualization results.

\subsection{Ablation on Loss Functions}

\textbf{The effectiveness of voxel-based losses} As given in Table~\ref{ablation_loss},  we verify the effectiveness of each training loss function in Equation 7.
From the results, we can observe the voxel-based losses, \textit{i.e}, $\mathcal{L}_\text{BCE}(\cdot)$ and $\mathcal{L}_\text{MSE}(\cdot)$, are important to train the proposed upsampling framework.
Specifically, the binary cross entropy loss $\mathcal{L}_\text{BCE}(\cdot)$ is calculated on occupancy classification probability $p_{c}$, and the mean squared error loss $\mathcal{L}_\text{MSE}(\cdot)$ is calculated on point density $\delta_{c}$.
Since the grid sampling relies on the predictions of occupancy classification probability and point density, imprecise predictions can result in inaccurate sampling, which leads to inferior performance in the upsampling task.

\begin{table}[t]
\begin{center}
\centering
\caption{Ablation studies on different loss functions. The experiments are conducted on PU1K dataset. The values of CD, HD, and P2F are multiplied by $10^{3}$.  }
\label{ablation_loss}
\resizebox{0.75\linewidth}{!}{
\begin{tabular}{p{2cm}<{\centering}|p{1cm}<{\centering}p{1cm}<{\centering}<{\centering}p{1cm}<{\centering}}
\toprule[1pt]
{Methods}&
{CD }&
{HD}&
{P2F}\\
\midrule[0.5pt]
w/o $\mathcal{L}^\text{S}_\text{CD}(\cdot)$ & 0.349&3.263&1.274\\
w/o $\mathcal{L}_\text{r}(\cdot)$ &0.351  &2.978 &1.212 \\
w/o $\mathcal{L}_\text{BCE}(\cdot)$&0.408  &3.229  &1.371 \\
w/o $\mathcal{L}_\text{MSE}(\cdot)$&0.389 &3.360 & 1.325\\
\midrule[0.5pt]
Full model&\textbf{0.338} &\textbf{2.694}&\textbf{1.183}\\
\bottomrule[1pt]
\end{tabular}}
\end{center}
\end{table}

\begin{table}[t]
    \resizebox{1\linewidth}{!}{
    \begin{tabular}{l|ccc|ccc}
    \toprule[1pt]
  \multirow{2}{*}{Methods}
  & \multicolumn{3}{c|}{{Sparse (1,024) input} }
  & \multicolumn{3}{c}{{Dense (2,048) input}}  \\
  & {CD}& { HD} & {P2F}& {CD} & { HD} & {P2F} \\
    \midrule[0.5pt]
    MPU &1.268&16.088&4.777  &0.861 & 11.799 &3.181 \\
    MPU (w/ GC) &1.129&13.973&4.028& 0.650& 9.341&2.653 \\
    \midrule[0.5pt]
    PU-GCN &1.036&12.026&3.938 &0.585 &7.757 & 2.499 \\
    PU-GCN  (w/ GC) &0.977&11.776& 3.678&0.570&8.181&2.393\\
  \midrule[0.5pt]
    Ours & 0.644&5.543&1.886&0.352 &2.809 & 1.283 \\
    Ours  (w/ GC) &0.617&5.383& 1.803&0.338&2.684&1.183\\
    \bottomrule[1pt]
    \end{tabular}}
    \caption{Wide applicability of latent geometric-consistent learning on different point cloud upsampling methods.
    }
    \label{GC_performance_w}
\end{table}

\textbf{Applicability of latent geometric-consistent learning.}
We further conduct an experiment to verify the effectiveness of our latent geometric-consistent learning method.
In this experiment, we applied the proposed scheme to two point cloud upsampling methods to explore its wide applicability.
The results in Table~\ref{GC_performance_w} show the advantage of our method in  improving other point cloud upsampling approaches by constraining the local surface patches in latent space.
Moreover, we notice our method also perform well on more sparse input point size, indicating its robustness to the different density distribution.

\begin{table}[t]
\begin{center}
\centering
    \resizebox{1\linewidth}{!}{
 \begin{tabular}{l|cc|cc|cc}
    \toprule[1pt]
  \multirow{2}{*}{Methods}
  &  \multicolumn{2}{c|}{0\%}
  &  \multicolumn{2}{c|}{1.0\%}
  &  \multicolumn{2}{c} { 2.0\%}  \\
  & {CD}& { HD}& {CD}& { HD}& {CD}& { HD} \\

    \midrule[0.5pt]
     PU-Net&1.157& 15.297&1.193& 16.203&1.343& 21.573\\
    MPU&0.861&11.799&0.882& 12.909&1.131& 16.598 \\
    PU-GAN &0.661&9.238 & 0.767&9.915&0.981 & 12.352 \\
    Dis-PU&0.731& 9.505&0.851 &12.272 & 1.112& 15.402\\
    PU-GCN&0.585&7.577&0.775  &9.812 & 1.082 & 13.159 \\
    MAFU&0.670&10.814&0.805  &11.323 & 1.071 & 12.836 \\
    Grad-PU&0.404&3.732 &0.602&5.896& 0.987& 9.369 \\
    \midrule[0.5pt]
    Ours&\textbf{0.338}&\textbf{2.694}&\textbf{0.554}&\textbf{5.399}&\textbf{0.943}&\textbf{9.238}\\
    \bottomrule[1pt]
    \end{tabular}
}
\caption{Quantitative comparisons ($\times 4$ upsampling rate) on PU1K dataset using noisy input point clouds with increasing noise levels. The input size of point clouds is 2,048. The values of CD, and HD are multiplied by $10^{3}$.}
    \label{noise}
\end{center}
\end{table}

\subsection{Noise Robustness Test}
To evaluate the effectiveness of PU-VoxelNet on noisy input, we conduct experiments on PU1K~\cite{Qian2021PUGCNPC} dataset by adding Gaussian noise.
The quantitative comparisons are presented in Table~\ref{noise}. From the results, we can see our method consistently achieves better performance than other counterparts.
We believe that the reason behind this is that our grid features are computed by averaging the point features within each grid cell.
As a result, the perturbations present in noisy inputs can be alleviated due to the averaging operation. This leads to our method exhibiting better robustness to noisy point clouds.

\begin{table}[t]

    \resizebox{1\linewidth}{!}{
    \begin{tabular}{l|p{1cm}<{\centering}p{1cm}<{\centering}p{1cm}<{\centering}cc}
    \toprule[1pt]
  \multirow{1}{*}{Methods}
  & {CD}& { HD} & {P2F}  & \multirow{1}{*}{ Size (Mb)} & \multirow{1}{*}{Time (ms)} \\
    \midrule[0.5pt]
    CDA&1.435 &16.815&5.467 &197.42& \textbf{22.32}\\
    GRNet &1.234 &13.441 & 4.873&292.63&44.91 \\
    VE-PCN & 0.855&9.605&3.116&117.29 &45.30 \\
    \midrule[0.5pt]
    Ours&\textbf{0.338}&\textbf{2.694}& \textbf{1.183}&\textbf{14.49}&44.56\\
    \bottomrule[1pt]
    \end{tabular}}
    \caption{Quantitative results on PU1K dataset with recent voxel-based point generative models. The inference time is testing on a Tesla P40 GPU.
    }
    \label{voxel-based-result}
\end{table}

\subsection{Comparisons to Voxel-based Models}
Recently, certain works~\cite{Lim2019ACD,xie2020grnet,wang2021voxel} employ voxel-based networks for point cloud generation or completion, which can also be directly used for upsampling of point cloud.
Here, we make a comparison with them in Table~\ref{voxel-based-result}, including CDA~\cite{Lim2019ACD}, GRNet~\cite{xie2020grnet}, and VE-PCN~\cite{wang2021voxel}.
As mentioned in the main text, these methods suffer from the inaccurate sampling problem caused by the imprecise predictions.
Besides, the downsampling operation used in VE-PCN~\cite{wang2021voxel} may not be suitable for upsampling, since the spatial information of sparse  points is initially inadequate, and GRNet~\cite{xie2020grnet} randomly samples coarse points and merely relies on several MLPs for denser point generation, which is also insufficient for generating new points.
As a result, these methods tend to have poor performance when it comes to the task of upsampling.

\begin{figure*}[ht!]
    \centering
    \includegraphics[scale=0.28]{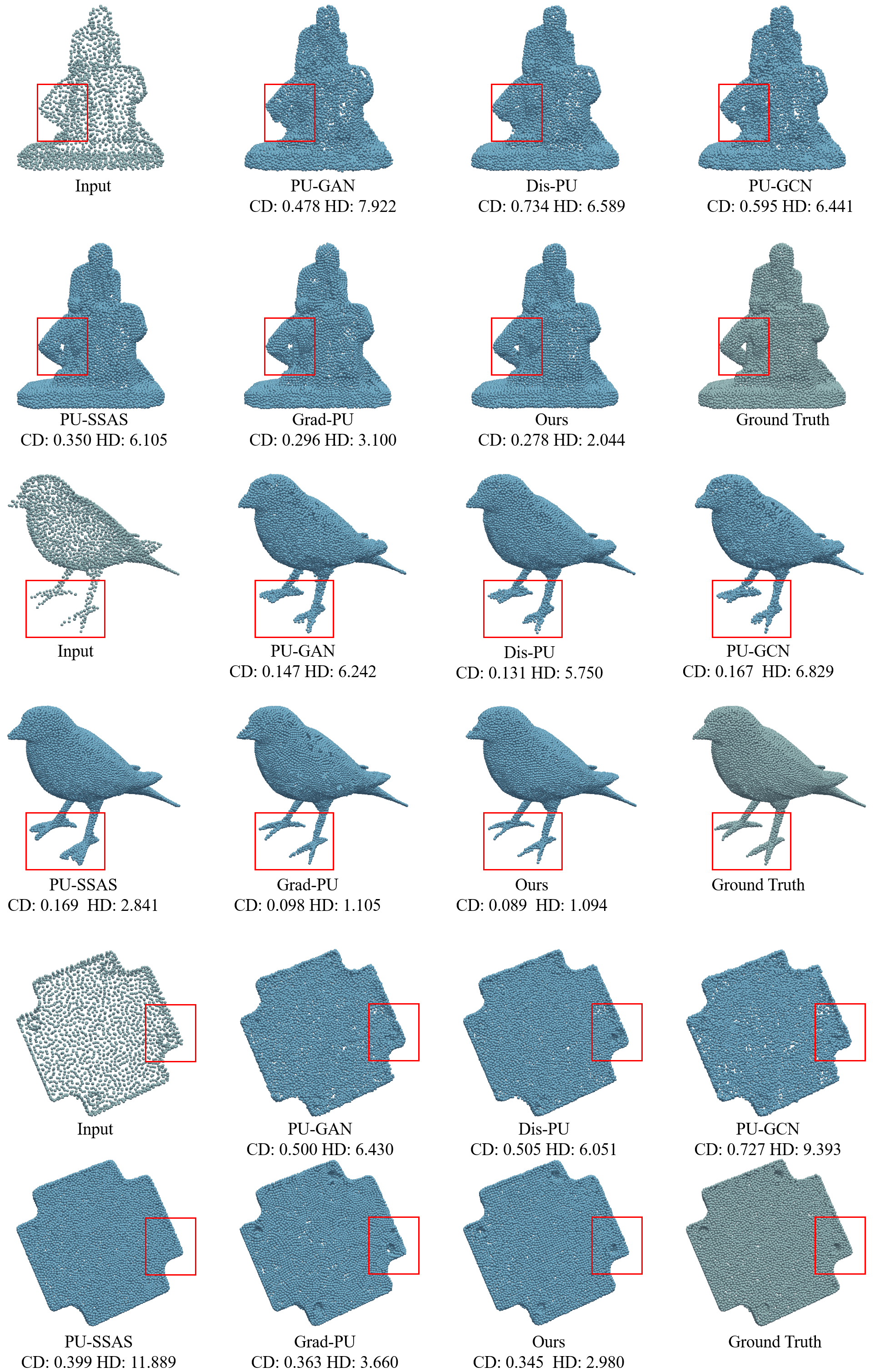}
    \caption{Qualitative results  ($\times 4$) on PU-GAN dataset with the input size of 2,048. Our method achieve high-fidelity surface approximation with more uniform point distribution. }
    \label{vis_pugan}
\end{figure*}

\subsection{More Visualization Results}
In the following, we provide more qualitative results on both synthetic and real-scanned data.

\textbf{Synthetic Data.} We have given some visualized results on PU1K dataset~\cite{Qian2021PUGCNPC} in the main text. Here, we also provide more qualitative results on PU-GAN dataset~\cite{Li2019PUGANAP}.
As shown in Fig.~\ref{vis_pugan}, we can clearly observe the proposed framework handles the complex geometry well and produces high-fidelity points with a uniform distribution, while other counterparts generate more outliers.

\begin{figure*}[t]
    \centering
    \includegraphics[scale=0.27]{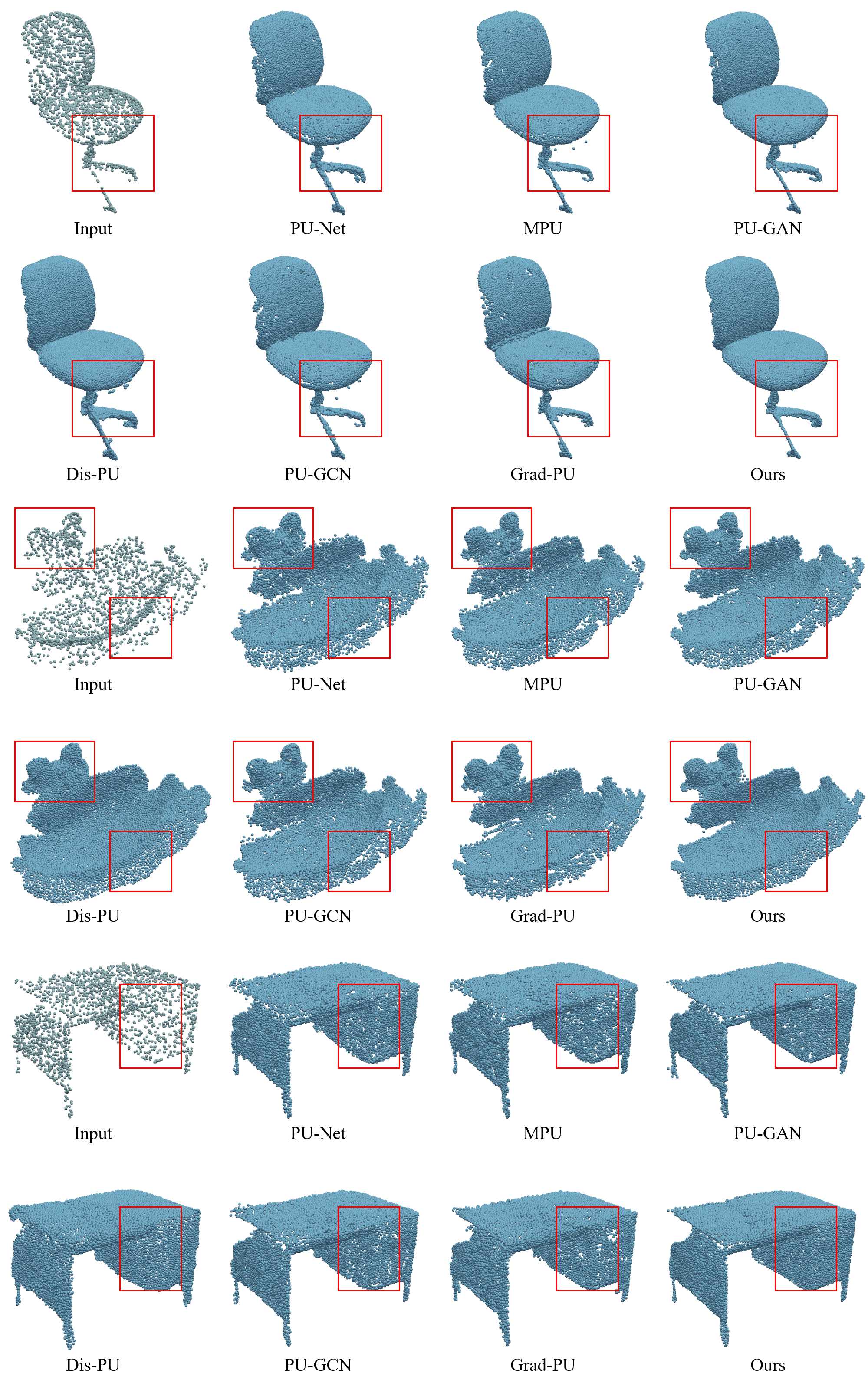}
    \caption{Qualitative results ($\times 4$) on real-scanned data from ScanObjectNN~\cite{Uy2019RevisitingPC} dataset.
    The input point clouds are sparse and non-uniform. After upsampling, we can generate more complete point clouds with a uniform point distribution.  }
    \label{vis_real}
\end{figure*}

\begin{figure*}[t]
    \centering
    \includegraphics[scale=0.32]{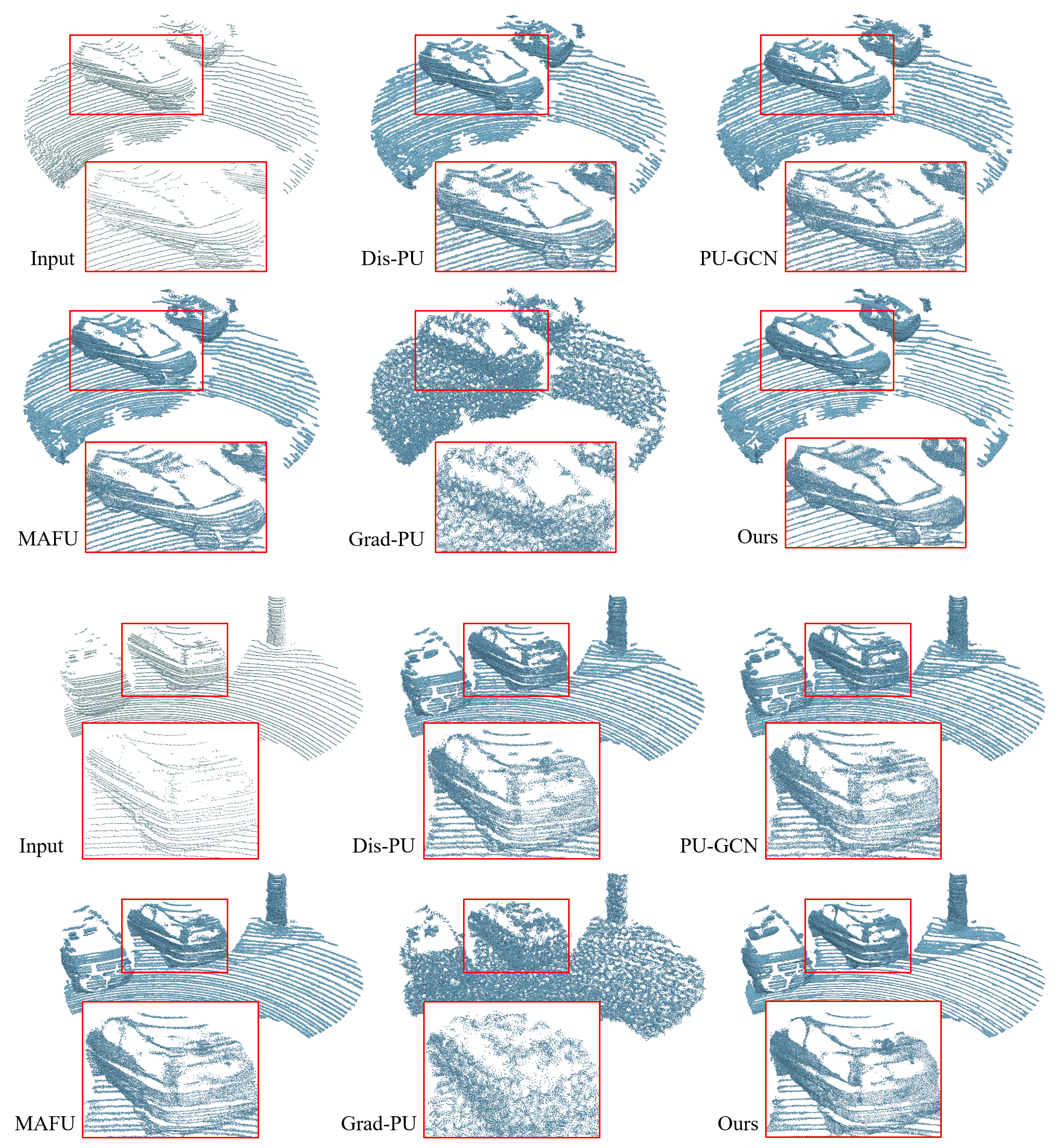}
    \caption{Qualitative results ($\times 4$) on real-scanned LiDAR data from KITTI~\cite{geiger2012we} dataset.  We can observe that the input point clouds are sparse, non-uniform  and incomplete. Compared with other counterparts, our method produces more fine-grained structure with fewer outliers in most cases. Moreover, when examining the upsampled results of Grad-PU~\cite{He_2023_CVPR}, we notice that its midpoint interpolation scheme may not work well on large-scale LiDAR data and its gradient-based updating fails to push the points into the underlying surface in such a scenario.  }
    \label{vis_4}
\end{figure*}

\textbf{Real-scanned Data.} As shown in Fig.~\ref{vis_real}, we give more results on real-scanned data from ScanObjectNN~\cite{Uy2019RevisitingPC} dataset.
In addition, we also provide several upsampled results on real-scanned LiDAR data from KITTI~\cite{geiger2012we} dataset in Fig.~\ref{vis_4}.
All the models are pre-trained on PU1K dataset. From the results, we can observe the real-scanned data is particularly sparse, non-uniform and incomplete.
In such scenarios, other methods tend to generate point distributions that are non-uniform and have more outliers.
In contrast, our method is able to produce more detailed structures while maintaining a faithful point distribution.

\end{document}